%% file: main.tex
\documentclass[10pt,twocolumn,letterpaper]{article}

\usepackage[pagenumbers]{cvpr} 

\usepackage{multirow}
\usepackage{stfloats} 
\usepackage{caption}

\newcommand{\mcircle}[1]{%
  \raisebox{1pt}{\textcircled{\raisebox{-0.9pt}{\textcolor{red}{#1}}}}%
}
\input{preamble}

\definecolor{cvprblue}{rgb}{0.21,0.49,0.74}
\usepackage[pagebackref,breaklinks,colorlinks,allcolors=cvprblue]{hyperref}

\def\paperID{} 
\def\confName{CVPR}
\def\confYear{2026}

\title{3D Space as a Scratchpad for Editable Text-to-Image Generation}

\author{
Oindrila Saha$^{1}$\thanks{Work done during internship at Adobe Research} \quad
Vojtech Krs$^2$ \quad
Radomir Mech$^2$ \\
Subhransu Maji$^{1}$ \quad
Matheus Gadelha$^{2}$ \quad
Kevin Blackburn-Matzen$^{2}$
\vspace{0.2cm} \\ 
$^1$University of Massachusetts Amherst \quad $^2$Adobe Research \\
{\tt\small \{osaha, smaji\}@umass.edu} \quad
{\tt\small \{vkrs, rmech, matzen, gadelha\}@adobe.com}
}

\begin{document}

\twocolumn[{%
\renewcommand\twocolumn[1][]{#1}%
\maketitle

\begin{center}
    \centering
    \captionsetup{type=figure} 
    \includegraphics[width=\linewidth]{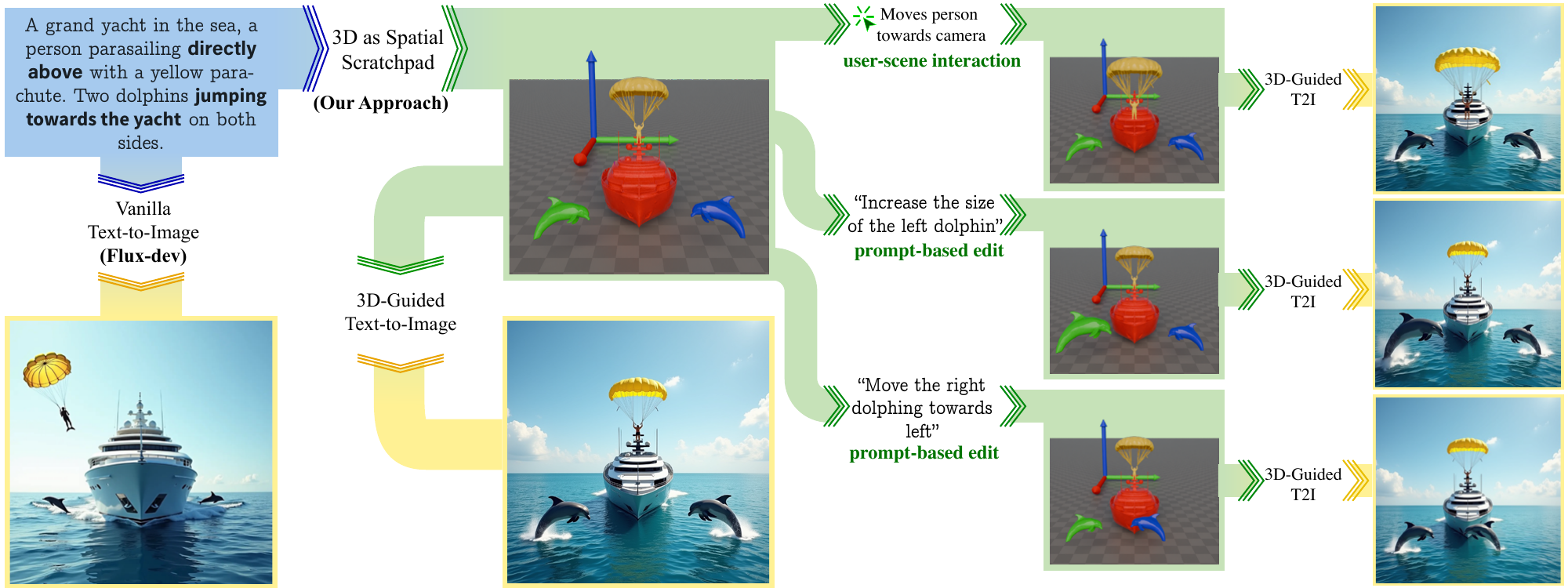}    
    \captionof{figure}{
   \textbf{Spatial scratchpads enable structured 3D reasoning for controllable image generation.}
Our method constructs a 3D spatial scratchpad from a text prompt, representing subjects as editable meshes with explicit geometry and spatial relations.
Users or language agents can manipulate this scene --- moving, resizing, or reorienting objects --- through either text-based or interactive edits.
The updated 3D configuration is re-rendered via a 3D-guided text-to-image pipeline, producing identity-preserving and spatially coherent images that remain faithful to the original prompt.
This demonstrates how 3D reasoning serves as an effective intermediate workspace linking linguistic intent and precise visual synthesis. 
    }
    \label{fig:teaser}
\end{center}%
}]

{
  \insert\footins{\noindent\footnotesize $^*$Work done during internship at Adobe Research.}
}

\input{sec/0_abstract}    
\input{sec/1_intro}
\input{sec/2_related_work}

\input{sec/3_method}

{
    \small
    \bibliographystyle{ieeenat_fullname}
    \bibliography{main}
}

\input{sec/X_suppl}

\end{document}

%% file: preamble.tex
\renewcommand{\paragraph}[1]{\vspace{.25em}\noindent\textbf{#1}.}



%% file: sec/0_abstract.tex
\begin{abstract}

Recent progress in large language models (LLMs) has shown that reasoning improves when intermediate thoughts are externalized into explicit workspaces, such as chain-of-thought traces or tool-augmented reasoning.
Yet, visual language models (VLMs) lack an analogous mechanism for spatial reasoning, limiting their ability to generate images that accurately reflect geometric relations, object identities, and compositional intent.
We introduce the concept of a spatial scratchpad -- a 3D reasoning substrate that bridges linguistic intent and image synthesis.
Given a text prompt, our framework parses subjects and background elements, instantiates them as editable 3D meshes, and employs agentic scene planning for placement, orientation, and viewpoint selection.
The resulting 3D arrangement is rendered back into the image domain with identity-preserving cues, enabling the VLM to generate spatially consistent and visually coherent outputs.
Unlike prior 2D layout-based methods, our approach supports intuitive 3D edits that propagate reliably into final images.
Empirically, it achieves a \textbf{32\%} improvement in text alignment on GenAI-Bench, demonstrating the benefit of explicit 3D reasoning for precise, controllable image generation.
Our results highlight a new paradigm for vision–language models that deliberate not only in language, but also in space.
Code and visualizations at \url{https://oindrilasaha.github.io/3DScratchpad/}.
\end{abstract}

%% file: sec/1_intro.tex
\section{Introduction}
\label{sec:intro}

Large language models have demonstrated remarkable abilities in reasoning when provided with mechanisms to externalize intermediate thoughts.
Early work in symbolic AI formalized such deliberation through logic, search, and planning frameworks \cite{newell1956logictheorist, mccarthy1960programs, FIKES1971189}.
Modern neural approaches have rediscovered this principle in linguistic form: by generating reasoning traces or invoking external tools, models can structure their computation over multiple steps.
Prompting strategies such as scratchpads \cite{nye2021scratchpad}, chain-of-thought reasoning \cite{wei2022chainofthought}, and least-to-most decomposition \cite{Zhou2022LeasttoMostPE} allow models to serialize latent computation into language.
Further, systems like ReAct \cite{yao2022react}, PAL \cite{gao2022pal}, and Toolformer \cite{schick2023toolformer} demonstrate that coupling language reasoning with external symbolic or procedural tools yields more accurate and interpretable behavior.
These advances underscore a key insight: reasoning improves when a model can write to and read from an explicit workspace.

Visual language models and text-to-image generators face a similar but unsolved challenge.
While they can describe and synthesize detailed imagery from text, they often struggle with precise spatial relations, consistent identities, and compositional coherence.
In language models, a scratchpad provides space to deliberate before answering; in image generation, no such spatial workspace exists.
Current layout- or region-conditioned approaches offer partial solutions—using masks, bounding boxes, or segmentation maps—but remain limited to coarse spatial hints~\cite{chen2024training,li2025seg2any}.
While recent work leveraged such systems to create imagery from intermediary 2D representations, a fundamental issue remains:
when dealing with spatial properties, restricting reasoning to planar entities significantly hinders the ability of
such systems to handle complex tasks.
The broader question is then how to equip visual generative models with a medium that allows them to ``think'' spatially, in \emph{three dimensions}, 
before creating the final image.

We introduce the concept of a \emph{spatial scratchpad} -- a 3D reasoning substrate that connects linguistic intent to visual synthesis.
Given a text prompt, an LLM first parses and identifies subjects and background elements.
Each subject is instantiated as a 3D mesh, which can then be positioned and oriented within a virtual scene by a collection of specialized agents responsible for placement, orientation, and camera selection.
The resulting 3D arrangement is rendered back into the image domain with identity-preserving and location cues, serving as a multiview spatial summary that a visual language model can interpret.
As we show in Figure~\ref{fig:teaser}, through this intermediate 3D reasoning representation, the model can align generated images more precisely with textual intent while also maintaining an editable format for the user. Our spatial scratchpad supports manual and prompt-based edits in space which can be reflected on to the generated images while maintaining the identity of all subjects and the background.

This formulation reframes 3D not as a rendering target, but as a reasoning substrate for image generation.
By explicitly grounding subjects in geometry, spatial scratchpads allow edits and constraints expressed in 3D space -- translations, rotations, rescaling -- to propagate reliably into final images.
Our agentic scene planning further decomposes control into interpretable components, enabling robust composition and user-guided editing.
Together, these elements achieve 3D-consistent, identity-preserving, and compositionally controllable image generation—capabilities largely absent in current 2D-centric systems.
Remarkably, our method improves scores by \textbf{32\%} on the challenging GenAI-Bench \emph{without requiring any additional training}.
These gains highlight the benefit of explicit 3D reasoning as an intermediate stage in visual generation, suggesting a path toward models that deliberate not just in language, but also in space. Our contributions are as follows:

\begin{itemize}
    \item We propose 3D space as a novel reasoning scratchpad for text-to-image generation.
    \item Through evaluation on complex compositional image generation benchmarks, we show that this reasoning modality improves text fidelity.
    \item We show that our designed scratchpad supports manual and text-based edits in 3D space that can be reliably and consistently reflected on to final generated images.
\end{itemize}

%% file: sec/2_related_work.tex
\section{Related Work}

\paragraph{Visual reasoning with LLMs}
LLMs have been proven to be an effective tool to understand and reason about diverse modalities of images and visual tasks. VisProg~\cite{gupta2023visual} and ViperGPT~\cite{suris2023vipergpt} showed that it was possible to use general purpose LLMs like ChatGPT out of the box to synthesize chains of visual tasks using existing code generation capabilities.  MM-ReAct~\cite{yang2023mm} combined ChatGPT with vision experts to do multi-modal planning in the loop with the LLM.  
Chain-of-Thought has been extended to the visual domain with systems such as Sketchpad~\cite{hu2024visual} which allow the LLM to generate drawings which it may use in future iterations. Whiteboard-of-Thought~\cite{menon2024whiteboard} specifically poses visual CoT in a framework where intermediate images are drawn with code.  Lofti~\etal~\cite{lotfi2024chain} focus on a broader set of Chain-of-Sketch problems that require multiple stages of visual reasoning to solve. In this work, we use LLMs to reason about renders of a 3D space, and estimate orientations of various subjects in images.

\begin{figure*}[h]
    \centering
\includegraphics[width=0.95\linewidth]{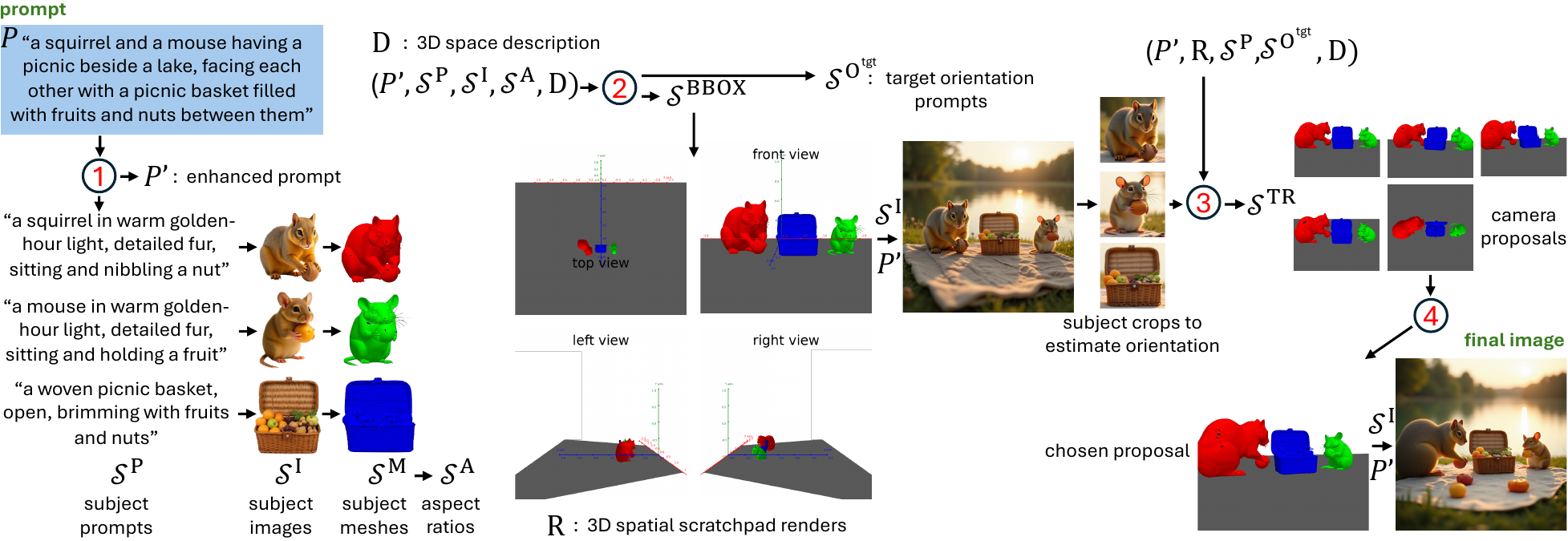}
    \caption{\textbf{Overview of a 3D spatial scratchpad.} Given an input prompt $P$ we illustrate how our method uses a 3D space as an underlying representation to generate an image that has superior alignment to the prompt. Agent \mcircle{1} is responsible for decomposing the input prompt into subjects and background. Agent \mcircle{2} provides 3D bounding boxes for each subject. We render the scratchpad and subsequently generate an image based on these placements which is then given to agent \mcircle{3} that adjusts transformations of the meshes. Finally, agent \mcircle{4} chooses the best camera viewpoint from a set of proposals to generate the final image. 
      \vspace{-5mm}}
    \label{fig:pipeline}
\end{figure*}

\paragraph{Image generation with LLM planning}
For generating images with complex compositional prompts, advanced reasoning is required to be faithful to the prompt. A direction of prior work show that reasoning and planning purely in the text modality through prompt enhancements can offer substantial improvements. RePrompt~\cite{chen2024reprompt} and PromptEnhancer~\cite{wang2025promptenhancer} train LLMs to produce detailed and structured prompts via reinforcement learning. Idea2Img~\cite{yang2023idea2img} is a training-free approach that uses off-the-shelf LLMs to iteratively rate generated images to improve prompts. Another active direction of research focuses on reasoning in the 2D space to generate images with regional decomposition. LayoutGPT~\cite{feng2023layoutgpt} can generate spatial layouts with an LLM in both 2D and indoor 3D domains.  LVD~\cite{lian2023llm} extends the layout problem to spatio-temporal layout for video. SLD~\cite{wu2024self} uses a LLM to produce and iteratively refine 2D bounding boxes based on feedback on generated images. RPG~\cite{yang2024mastering} exploits chain-of-thought prompting to improve regional planning with LLMs. MUSES~\cite{ding2025muses} also uses LLMs for planning 2D bounding boxes, followed by depth prediction of each box. In contrast, to the best of our knowledge, our method is the first approach to introduce LLM based planning purely in a 3D space to achieve text-to-image generation.

\paragraph{Conditional image generation} Several mechanisms for controlling image generation are in frequent use to date.  Such mechanisms can control for either image structure or subject identity. ControlNet~\cite{zhang2023addingconditionalcontroltexttoimage} and T2I-Adapter~\cite{mou2024t2i} show that pre-trained diffusion models can be controlled with diverse types of controls including pose, depth, and edges. Other works such as GLIGEN~\cite{li2023gligen}, BoxDiff~\cite{xie2023boxdiff}, and InstanceDiffusion~\cite{wang2024instancediffusion} devise methods to condition image generation on 2D bounding boxes. Several methods exist for encoding subject identity control.  DreamBooth~\cite{ruiz2023dreambooth} and Textual Inversion~\cite{gal2022image} learn per-subject identities from a set of reference images.  OminiControl~\cite{tan2025ominicontrol} accomplishes identity control for single subjects with a non-spatially aligned positional encoding.  SIGMA-Gen~\cite{saha2025sigma} builds on this framework by allowing control over multiple subject identities in a single image generation while also offering varying degrees of control precision, from 2D bounding box to full segmentation and with optional depth control. In this work, we adopt SIGMA-Gen for the image generation step so as to be able to control structure and identities of multiple subjects in a single denoising loop.

\paragraph{3D scene generation and layout} Due to the success of spatial reasoning from off-the-shelf LLMs, several works focus on generating structured 3D scenes by placing assets.  Yang~\etal~\cite{yangoptiscene} present OptiScene, an LLM aligned on a novel dataset for 3D home interior layout data using direct preference optimization. SceneFormer~\cite{wang2021sceneformer} similarly adds one object at a time to home interior scenes via model selection, placement, and rotation tasks. SceneCraft~\cite{hu2024scenecraft} presents a Blender-based framework wherein assets are retrieved, placed in a 3D scene, and refined with an LLM. 
Scenethesis~\cite{ling2025scenethesis} and CAST~\cite{yao2025cast} use an image to lift it up to 3D scenes using LLM planning. In contrast, we propose a framework with 3D as an underlying representation to generate images of diverse domains and show that this approach improves text fidelity of image generation.

%% file: sec/3_method.tex
\section{Method}
Our objective is to generate an image ${I}$ that follows the description of text prompt ${P}$.
For cases where ${P}$ requires complex compositions, directly generating $I$ from ${P}$ using a
generative model (\eg a diffusion model) is very challenging.
Thus, using the prompt ${P}$, we create an intermediate 3D scratchpad representation.
To begin with, the 3D scratchpad is an empty scene with a ground plane and fixed bounds in $X$, $Y$, and $Z$ axes.
We set a fixed lighting to the environment.
Multiple subjects can be instantiated, arranged and oriented in the scratchpad so as to generate a 2D image ${I}$ that faithfully adheres to the given user prompt.

This representation enables us to perform 3D-aware image editing either manually or with user prompts while maintaining 3D consistency and individual subject identities. The task can be accomplished when broken down into several sub-tasks: \mcircle{1} identifying and instantiating subjects, \mcircle{2} planning placements in 3D space, \mcircle{3} adjusting 3D transformations of each subject, and \mcircle{4} placing the camera. We employ the use of an agentic framework with one or more LLM agents to accomplish each of the sub-tasks. Finally, we use multi-subject identity and depth controlled image generation to create image ${I}$ conditioned on the 3D spatial scratchpad. Our pipeline is illustrated in Figure~\ref{fig:pipeline}.

\subsection{Identifying and instantiating subjects - \mcircle{1}}
Given a prompt ${P}$ we can decompose it into a set of $n$ subjects $s_1, s_2, \dots s_n \in \mathcal{S}$ plus a background. We use a LLM agent that extracts individual subjects and forms descriptions $s^P_1, s^P_2, \dots s^P_n \in \mathcal{S}^\mathrm{P}$ of each from the input prompt. We also ask an LLM agent to create an enhanced prompt $P'$ based on the above reasoning. Each of these descriptions can then be used to create 3D objects either by 1) text-to-3D models, or 2) text-to-image followed by image-to-3D models. We choose the second path so as to ensure identity control in the final step of image generation. We show in Figure~\ref{fig:d2ivssigma} that image-based identity control improves prompt adherence. The text-to-image step provides us with identity images for each subject $s^I_1, s^I_2, \dots s^I_n \in \mathcal{S}^\mathrm{I}$. In succession, image-to-3D over $\mathcal{S}^\mathrm{I}$ provides us 3D meshes $\mathcal{S}^M$ of each subject. For efficiency, we do not include texture generation of the individual 3D subjects. Rather we assign fixed colors to each mesh and associate these colors to each subject in our LLM prompts.\footnote{Note that the appearance of each subject in the final image is controlled only by the subject identity images $\mathcal{S}^I$ and depth.}

\subsection{Planning placements in 3D space - \mcircle{2}}
We describe this 3D space in text ($D$) with respect to a front camera $C^{\mathrm{front}}$ and provide it to every subsequent LLM agent. The description contains information about axes, position of ground plane and bounds. Given the subject meshes $\mathcal{S}^\mathrm{M}$ and their descriptions $\mathcal{S}^\mathrm{P}$ we can use a LLM planning agent to place these subjects based on prompt ${P'}$. 

We use LLMs to generate 3D bounding boxes of each subject given a description of a blank 3D space. To obtain reliable 3D bounding boxes from a LLM, we equip the LLM with multiple modes of relevant information. We provide the global prompt $P'$, the individual subject prompts $\mathcal{S}^\mathrm{P}$, and the 3D aspect ratios of each subject $\mathcal{S}^\mathrm{A}$. The LLM agent provides
\begin{align}
    \mathcal{{S}}^\mathrm{BBOX} = \mathrm{BboxPlanner}(P', \mathcal{S}^\mathrm{P}, \mathcal{S}^\mathrm{I}, \mathcal{S}^\mathrm{A}, D)
\end{align}

The global prompt ${P'}$ provides information about relative placement, the subject prompts $\mathcal{S}^\mathrm{P}$ provides context about each of the subjects to be placed, and the aspect ratios $\mathcal{S}^\mathrm{A}$ aid in planning ratios of the 3D bounding box that the LLM generates. 
Using these generated bounding boxes $\mathcal{{S}}^\mathrm{BBOX}$ we place the subject meshes $\mathcal{S}^\mathrm{M}$ in the 3D scene. We also instruct this LLM agent to provide a high level orientation description for each subject ($\mathcal{S}^\mathrm{O_{tgt}}$) in natural language such as ``facing the camera", ``lying flat" and so on, which is used in the following step.

\subsection{Adjusting the 3D scratchpad - \mcircle{3}}
Given initial placements of each subject mesh, we now move to adjusting their rotations, translations and scales. We find that understanding 3D rotations is more challenging for LLMs as compared to planning 3D placements. We show in the supplementary that simply using the renders of the current 3D scene to obtain rotation suggestions from an LLM is not reliable. For this reason, we divide this task among two LLM agents. The first agent, $\mathrm{OrientationEstimator}$, identifies the current absolute orientation of each subject from an image generated by conditioning on depth of the current 3D scene rendered from front camera $C^{\mathrm{front}}$ and identities $\mathcal{S}^\mathrm{I}$ (see Section~\ref{sec:sigmagen}). We discuss the effectiveness of this strategy in the supplementary. The second agent, $\mathrm{TransformPlanner}$, uses the target and estimated orientations along with 3D space description ($D$) to provide transformation suggestions of each subject. 

The $\mathrm{OrientationEstimator}$ agent is provided crops around each subject in the generated image conditioned on the 3D scene and identities, to estimate their orientation. We show in the supplementary that providing the full generated image along with the subject descriptions $\mathcal{S}^\mathrm{P}$ fails. We opt to use crops from the generated image to estimate orientation for each subject. We do not use the 3D scratchpad renders for this step due to our meshes being texture-less. We obtain $\mathcal{S}^\mathrm{O_{est}}$ from this agent which signifies the estimated absolute orientations in unconstrained natural language of each subject such as ``facing the camera", ``lying flat" etc.

Next we render our current 3D space from various cameras including front ($C^{\mathrm{front}}$), left, right, and top (see Figure~\ref{fig:pipeline} for examples). These renders ($R$) are provided to the $\mathrm{TransformPlanner}$ agent to understand the current state of the populated 3D scene. This agent provides transformation suggestions
\begin{align}
    \mathcal{{S}}^\mathrm{TR} = \mathrm{TransformPlanner}(P', R, \mathcal{S}^\mathrm{O_{est}}, \mathcal{S}^\mathrm{O_{tgt}}, \mathcal{S}^\mathrm{P}, D)
\end{align}
The agent uses the target absolute orientation $\mathcal{S}^\mathrm{O_{tgt}}$ and the estimated absolute orientation $\mathcal{S}^\mathrm{O_{est}}$ to suggest 3D rotations for each subject (see Figure~\ref{fig:pipeline}). Even though rotation suggestion is the primary task, we also instruct this agent to provide translation/scaling adjustments if required. Using these updated transformations we re-orient and adjust position/scales of each of the subjects.

\subsection{Placing camera - \mcircle{4}}
With this arranged and oriented 3D scene we can now generate our final image. For this we need to finalize a camera position. Using the front camera by default may lead to lack of aesthetic appeal or prompt adherence (see Fig~\ref{fig:scratch_progress}). Hence we introduce an LLM agent $\mathrm{CameraPicker}$ to decide the final camera. We find that simply instructing a LLM for camera placement and rotation coordinates produces unreliable outputs due to the problem being too abstract. To solve this, we render the scene from five proposal views constructed so as to contain all the subjects in the frame. Finally, the $\mathrm{CameraPicker}$ agent chooses one of these proposal renders ($C$) based on the prompt $P$.

\begin{figure}[h]
    \centering
\includegraphics[width=1.0\linewidth]{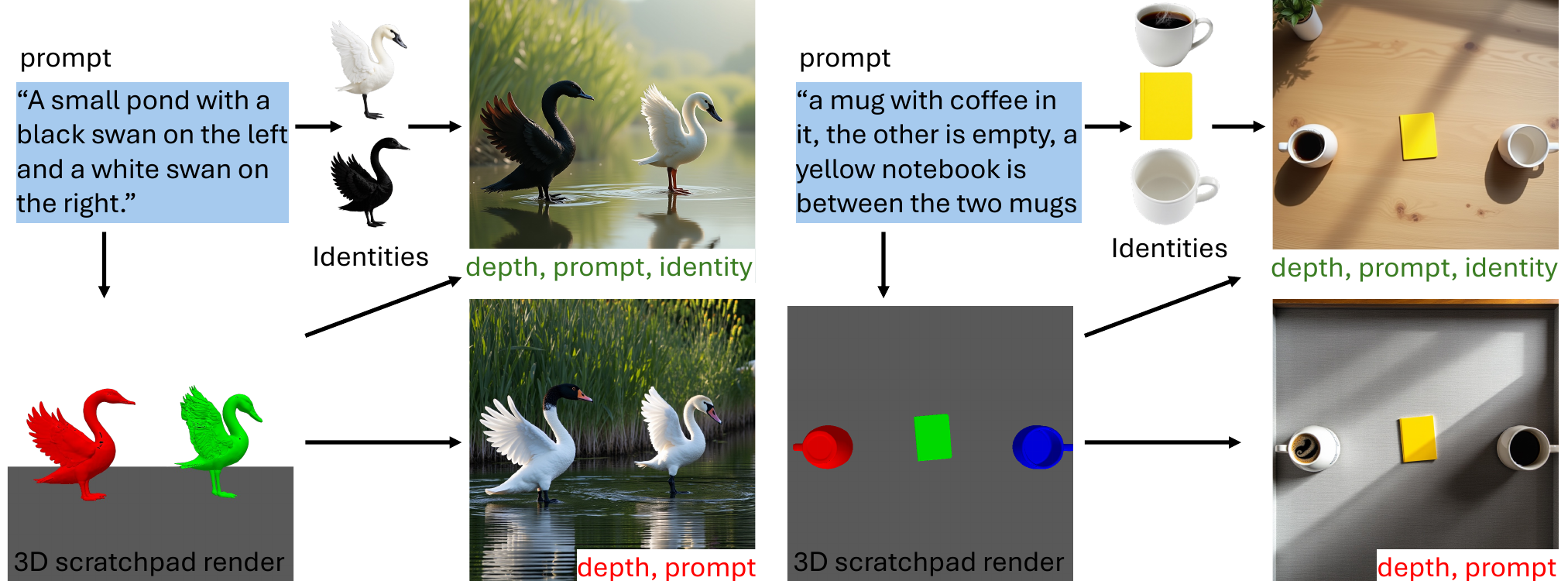}
    \caption{\textbf{Identity preservation improves prompt adherence.} We show that complex planning among multiple subjects even when guided with only depth and prompt can lead to loss of text alignment. In contrast, we opt to use depth, prompt, and identity to generate the images thus preserving prompt adherence.
\vspace{-5mm}}
    \label{fig:d2ivssigma}
\end{figure}

\paragraph{3D scene to image generation}
\label{sec:sigmagen}
We use depth as a condition to relay 3D information to text-to-image generation. With the depth, identity images $S^\mathrm{I}$ and enhanced prompt $P'$ we can obtain $I$ using a controllable identity preserved image generation method. In Figure~\ref{fig:d2ivssigma}, we show the importance of identity preserved image generation to create images that are faithful to the text prompt. For image editing we introduce an intermediate step of removing the subject(s) to be edited in the image using an image inpainting method, after which we can proceed with subject insertion at updated location/orientation(s).

\subsection{3D-aware image editing}
Our 3D representation equips us to perform identity and 3D consistent edits on the generated image. This can be approached through manual adjustments in the 3D space such as scaling, translation, and rotation. Another approach is to use an LLM agent to convert user-editing instructions to 3D transformations. The LLM agent $\mathrm{SubjectEditor}$ provides
\begin{align}
    \mathcal{{S}}^\mathrm{TR} = \mathrm{SubjectEditor}(E, I, C, \mathcal{S}^\mathrm{P}, \mathcal{{S}}^\mathrm{BBOX}, D)
\end{align}
The image and $I$ provides context to identify which subject(s) the user is referring to as well as for understanding the edit. The camera used to render depth for the image $C$ provides context to be able to translate image-level instructions to the world coordinate space. $\mathcal{{S}}^\mathrm{BBOX}$ provides exact size and position information to plan updates. $E$ is the user edit instruction. As usual we provide the subject prompts $\mathcal{S}^\mathrm{P}$ and the 3D space description $D$.
After applying these transformations $\mathcal{{S}}^\mathrm{TR}$, we can update our generated image using the method described in the section above, while preserving background and subject identities.

\section{Experiments}

\paragraph{Implementation details}
We use GPT-5~\cite{openai_gpt5_2025} as the LLM for each of the planning agents and GPT-4o~\cite{openai_gpt4o_2024} to extract formatted texts from the GPT-5 answers. We use Flux.1 [dev]~\cite{flux1dev2024} for generating images from each of the subject prompts, and Hunyuan-3D 2.5~\cite{lai2025hunyuan3d} for generating 3D meshes given the subject images. We use Pytorch-3D~\cite{ravi2020pytorch3d} for rendering our 3D scratchpad. To render the images we use a gray ground plane along with rulers drawn for the X, Y, and Z axes. We include more details in the supplementary. Finally, for identity preserved image generation we use SIGMA-Gen~\cite{saha2025sigma}. This choice is motivated by SIGMA-Gen's capability to handle identities of multiple subject as well as depth control in unison. This improves image quality compared to iteratively inserting subjects using subject insertion methods like InsertAnything~\cite{song2025insert}. However, we show in Table~\ref{tab:sigmavsia} that text alignment on using Insert-Anything with our method remains better than other reasoning modality baselines. This points to the fact that our method is compatible with future developments in multi-subject identity preserved image generation.
For editing the generated image, we perform removal of the subject to be edited using ObjectClear~\cite{zhao2025objectclear}. Then, for background preserved subject insertion, we use SIGMA-Gen equipped with latent blending~\cite{avrahami2022blended}. We use a dilated bounding box of the subject as the editing mask to allow better harmonization and shadows, while also providing the depth to guide precise structure.

\paragraph{Evaluation}
To evaluate our method on complex compositional prompts, we use the GenAI-Bench~\cite{li2024genai} and CompoundPrompts~\cite{sella2025instancegen} datasets. We use the advanced subject of GenAI-Bench, which contains 870 prompts. Additionally, GenAI-Bench also contains labels of the type of reasoning involved for a particular prompt. This includes: ``counting", ``comparison", ``differentiation", ``negation", and ``universal". We use the full CompoundPrompts dataset, which includes 540 prompts. Both of these datasets require advanced reasoning for image generation. To evaluate adherence to text we use VQAScore~\cite{lin2024evaluating} which uses a VQA model to find the probability of ``Yes" given the image and a question which asks whether the image is faithful to the input prompt. For evaluating image quality, we use Q-Align~\cite{wu2023q} that uses a large multimodal model for assessing quality. We also include evaluation on relatively simpler T2I-CompBench~\cite{huang2023t2i} dataset in the supplementary.

\paragraph{Baselines}
For improving prompt adherence of image generation for complex compositional prompts, advanced reasoning is required. This reasoning could be in only the text space, the 2D image space, or a 3D space like ours. For text-only reasoning, we choose Idea2Img~\cite{yang2023idea2img}, a strong baseline that performs prompt enhancement based on iterative feedback obtained on images generated using a LLM. We update the LLM used by the original method with GPT-5 and the image generation model with Flux.1[dev] for fair comparison to our method. For a baseline that employs reasoning over a 2D space we equip the highly effective RPG~\cite{yang2024mastering} method with GPT-5. By default, RPG uses SDXL~\cite{podell2023sdxl} as the base model, but we also equip it with Flux.1[dev] with regional prompting~\cite{chen2024training} for fair comparison. 

\begin{figure}[h]
    \centering
\includegraphics[width=1.0\linewidth]{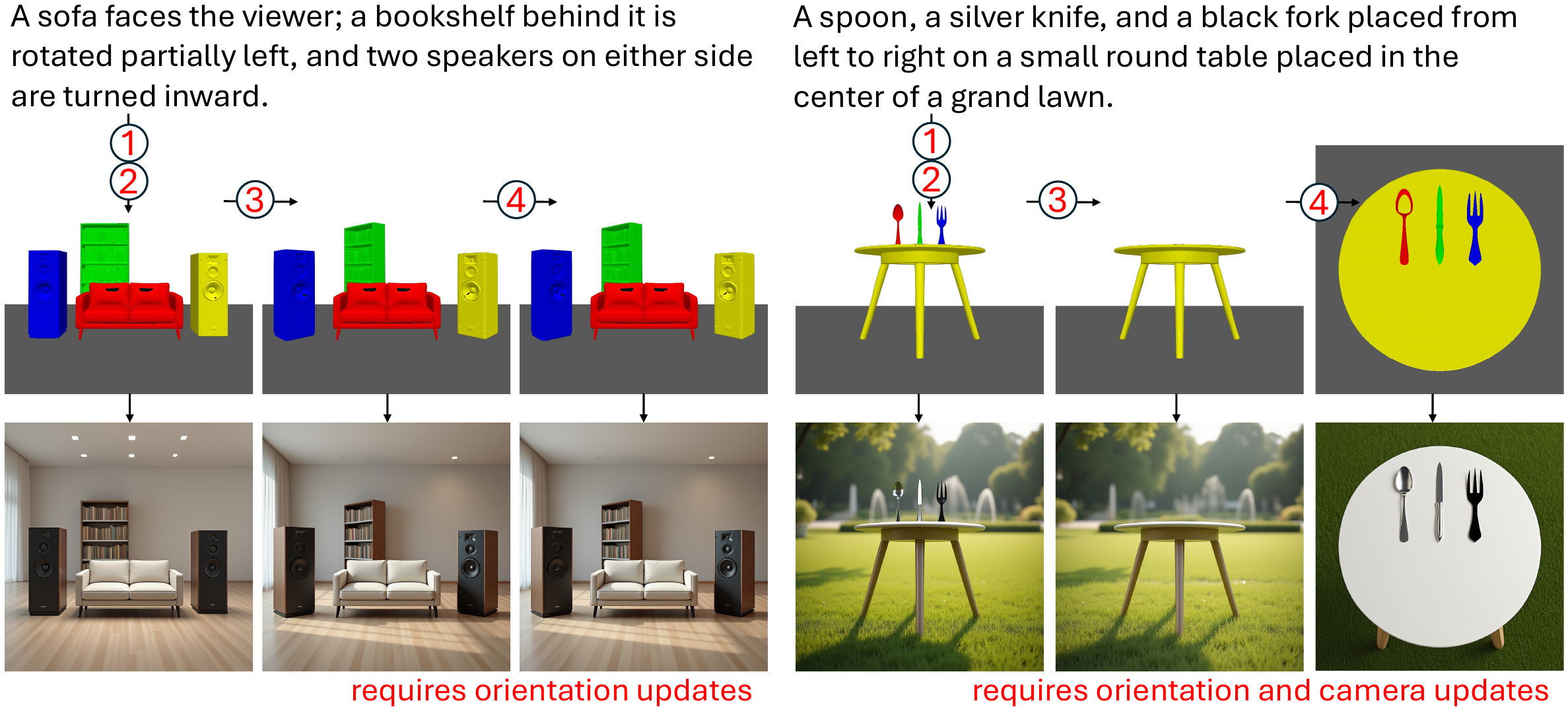}
    \caption{\textbf{Progression of 3D spatial scratchpad.}
We show examples where the planning conducted by agents \mcircle{3} and \mcircle{4} are crucial for being faithful to the text. In the first example, only \mcircle{3} is useful, while on the other both \mcircle{3} and \mcircle{4} are required.\vspace{-3mm}}
    \label{fig:scratch_progress}
\end{figure}

\begin{figure*}[h]
    \centering
\includegraphics[width=1.0\linewidth]{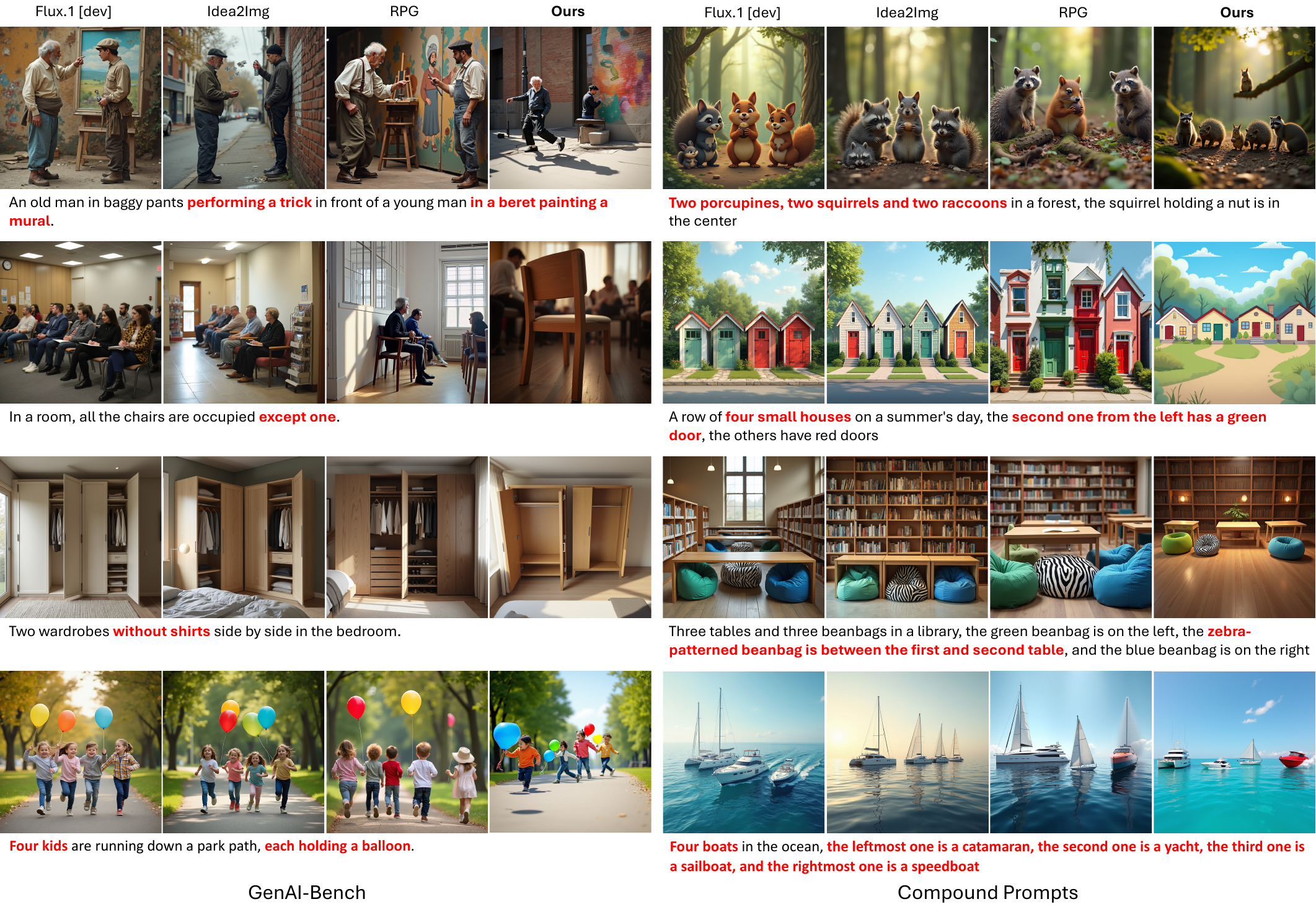}
\vspace{-7mm}
    \caption{
    \textbf{Qualitative comparison on GenAI-Bench and CompoundPrompts datasets.} We show examples where our method correctly captures the prompt, while the baseline methods fail. The GenAI-Bench dataset examples capture position, action, negation, and attribute planning. CompoundPrompts dataset examples show accurate attribute planning combined with spatial planning for many subjects.
\vspace{-2mm}}
    \label{fig:qualitative}
\end{figure*}

\begin{table}[]
\centering

\scriptsize
\begin{tabular}{l|c|cc|cc}
\toprule
\multirow{3}{*}{Method} &
\multirow{3}{*}{
\begin{tabular}{@{}c@{}}
Reasoning \\
modality
\end{tabular}
} &
\multicolumn{2}{c|}{GenAI-Bench} &
\multicolumn{2}{c}{CompoundPrompts} \\
& &
\begin{tabular}{@{}c@{}}Text \\ Alignment\end{tabular} &
\begin{tabular}{@{}c@{}}Image \\ Quality\end{tabular} &
\begin{tabular}{@{}c@{}}Text \\ Alignment\end{tabular} &
\begin{tabular}{@{}c@{}}Image \\ Quality\end{tabular} \\ \hline
Flux       & -    &    0.63      &     4.75     &    0.85      &     \textbf{4.84}     \\
Idea2Img           & text &     0.80     &    4.76      &    0.87      &      4.76    \\
RPG - SDXL       & 2D   &    0.60      &     4.65     &     0.71     &     4.66     \\
RPG - Flux   & 2D   &     0.71     &    \textbf{4.81}      &     0.84     &     4.83     \\
Ours       & 3D   &     \textbf{0.83}     &     \textbf{4.81}     &     \textbf{0.91}     &    \textbf{4.84}     \\
\bottomrule
\end{tabular}
\caption{\textbf{Quantitative comparison with baselines.} We show that using a 3D space as a reasoning scratchpad improves text alignment compared to using only text or a 2D space as a reasoning modality. We also maintain or improve in terms of image quality.}
\label{tab:main_table}
\end{table}

\begin{table}[]
\centering
\footnotesize
\begin{tabular}{l|cc}
\toprule
\multirow{2}{*}{Method}         & Text  & Image \\
& Alignment & Quality\\\hline
Idea2img (maxiter=1)        &     0.75     &    4.76      \\
Ours     &     0.83     &    \textbf{4.81}      \\   
Idea2img (full)  &     0.80     &      4.76    \\
Ours + Idea2img (full)  &   \textbf{0.85}    &     4.80     \\
\bottomrule
\end{tabular}
\caption{\textbf{Impact of prompt enhancement.} On GenAI-Bench, we show that using the iteratively improved prompt from Idea2img works complementarily to our approach and offers further improvement. We also show that using the Idea2img method but only for a single iteration reduces their performance significantly. We also maintain image quality with the different prompts.\vspace{-5mm}}
\label{tab:prompt_enhancement}
\end{table}

\begin{figure*}[h]
    \centering
\includegraphics[width=1.0\linewidth]{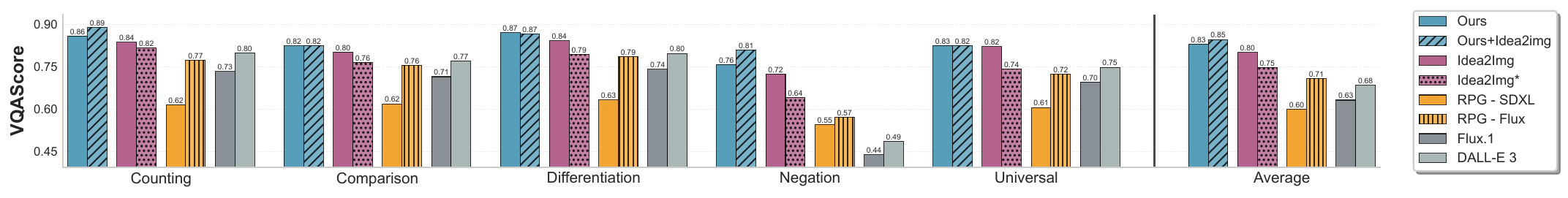}
    \caption{
\textbf{Comparison of text-to-image performance in GenAI-Bench.}
We evaluate our spatial scratchpad framework against state-of-the-art text-to-image systems on five reasoning categories—\emph{Counting}, \emph{Comparison}, \emph{Differentiation}, \emph{Negation}, and \emph{Universal}—as well as the overall average.
Scores reflect VQAScore accuracy (higher is better).
Across all categories, our single-iteration variants (\textbf{Ours+SIGMAGen} and \textbf{Ours+Idea2Img}) consistently outperform prior models.
Notably, \textbf{Idea2Img*} here denotes the single-iteration version of Idea2Img (without multi-step refinement), allowing a fair comparison with our own single-pass inference.
\vspace{-5mm}    }
    \label{fig:plot}
\end{figure*}

\begin{figure*}[h]
    \centering
\includegraphics[width=0.95\linewidth]{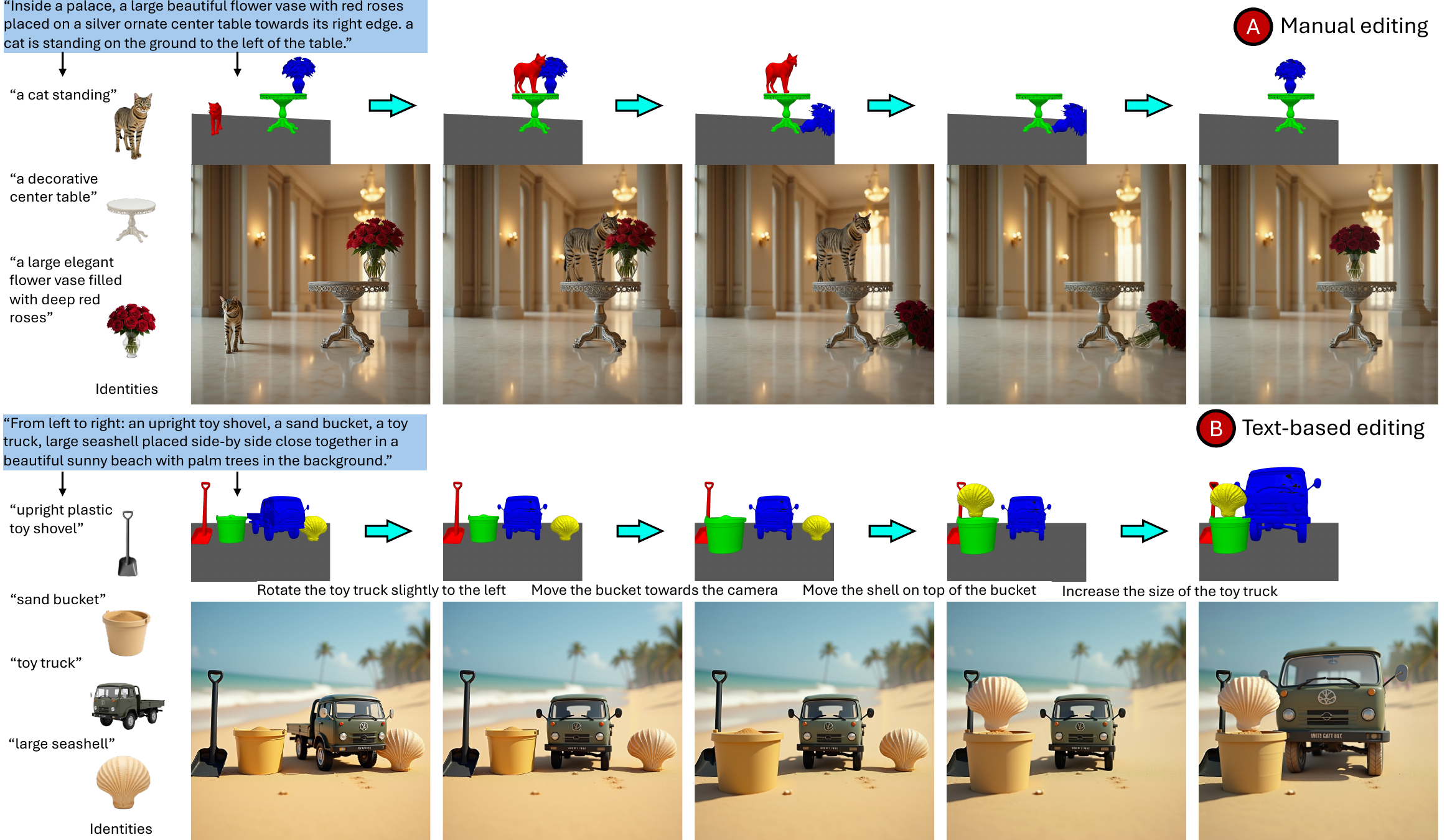}
    \caption{\textbf{3D as a spatial scratchpad enables consistent image editing.} Given the input prompt, we show the subjects instantiated, the scratchpad created and the subsequent image generated in the first column. In the succeeding columns, we show how edits made either through a) manual editing, or b) text-based editing to the scratchpad can be consistently reflected on to the final image while conserving identities of subjects and the background. Each column shows a progressive edit made over the state in the previous column.
\vspace{-3mm}}
    \label{fig:editing}
\end{figure*}

\begin{table}[]
\centering
\footnotesize
\begin{tabular}{l|cc}
\toprule
\multirow{2}{*}{Agents used}         & Text  & Image \\
& Alignment & Quality\\\hline
\mcircle{1} + \mcircle{2}      &    0.821      &      \textbf{4.81}    \\
\mcircle{1} + \mcircle{2} + \mcircle{3}      &       0.824   &    4.80     \\
\mcircle{1} + \mcircle{2} + \mcircle{3} + \mcircle{4} &     \textbf{0.830}     &     \textbf{4.81}     \\
\bottomrule
\end{tabular}
\caption{\textbf{Impact of each agent.} On GenAI-Bench, we show that both the \mcircle{3} and \mcircle{4} agents offer progressive improvement in text alignment while image quality stays stable across the ablations.}
\label{tab:agent_ablation}
\end{table}

\begin{table}[]
\centering
\footnotesize
\begin{tabular}{l|cc}
\toprule
\multirow{2}{*}{Scratchpad render choice}         & Text  & Image \\
& Alignment & Quality\\\hline
ground+back plane         &    0.821      &   4.81       \\
ground plane      &     0.821     &     \textbf{4.82}     \\
ground+back plane, rulers &     0.829     &   \textbf{4.82}       \\
ground plane, rulers      &    \textbf{0.830}      &   4.81       \\
grid, rulers              &     0.826     &   4.81      \\
\bottomrule
\end{tabular}
\caption{\textbf{Impact of design choice of renders.} On GenAI-Bench, we show that adding rulers to the 3D scratchpad's rendered images improves text alignment. With rulers the performance remains comparable. Image quality stays stable over all the choices.}
\label{tab:design_ablation}
\end{table}

\begin{table}[]
\centering
\footnotesize
\begin{tabular}{l|cc}
\toprule
\multirow{2}{*}{Image generation method}         & Text  & Image \\
& Alignment & Quality\\\hline
Iterative Insert-Anything*      &   0.81       &     4.60     \\
SIGMA-GEN      &    \textbf{0.83}      &      \textbf{4.81}    \\
\bottomrule
\end{tabular}
\caption{\textbf{Impact of identity preserved image generation method.} On GenAI-Bench, we show that Iterative Insert-Anything* (row 1), which refers to iterative insertion of subjects using Insert-Anything + depth ControlNet, leads to reduction of quality, however text alignment stays higher than other baselines.\vspace{-5mm}}
\label{tab:sigmavsia}
\end{table}

\paragraph{Comparison to baselines}
We show in Table~\ref{tab:main_table} that our method outperforms the baselines that use text or 2D as a reasoning modality. Both our method and RPG use a LLM to rephrase and form the text prompt(s) for the final image generation. It should be noted that Idea2Img uses a multi-step (3 iterations by default) prompt enhancement technique based on feedback on images generated in each iteration over 3 prompt variations. Finally the prompt that produced the image with the best score is selected. Since good-quality prompts are crucial in guiding diffusion models like Flux, Idea2Img offers significant improvement in text alignment over the baseline Flux.1 [dev] inferenced with original prompt (row 1). However, we show that using a 3D spatial scratchpad ensures improvement in terms of prompt adherence even with a single-iteration prompt enhancement. \textit{It should also be noted that none of the baselines support consistent editability of generated images.} Our method is also comparable or better than all baselines in the image quality of the generated image. We show qualitative examples of our method compared to baselines in Figure~\ref{fig:qualitative}. 

In Table~\ref{tab:prompt_enhancement} we test our method by starting with the best prompts chosen by Idea2img. We observe that this setup is complementary to ours and offers an extra boost (row 4). We also test Idea2img with only a single iteration of prompt enhancement (row 1) and note that this leads to considerable reduction in text alignment.

In Figure~\ref{fig:plot}, we show a breakdown of the performance of the baselines and our method over the various reasoning models in the GenAI-Bench dataset. Our method improves over the baselines in each of the reasoning modes considered. Since diffusion models struggle in handling negations, our method offers the highest improvement in that aspect. In this figure, we also include and improve over DALL-E-3~\cite{openai2023dalle3} which is the best performing image generation model as benchmarked by GenAI-Bench.

\paragraph{Importance of each agent}
The first and second agent are essential for instantiating subjects and placing them in appropriate positions. In Table~\ref{tab:agent_ablation}, we show that adding the third and fourth agents offer progressive improvement in text alignment. However, since the GenAI-Bench dataset does not involve many prompts that require accurate orientation or camera planning, which are the main contributions of the third and fourth agent, the differences observed are minimal. In Figure~\ref{fig:scratch_progress}, we show cases where the third and fourth agent help in maintaining prompt adherence. 

\paragraph{Impact of design choices}
We show in Table~\ref{tab:design_ablation} that our performance remains stable over choices of rendering design, however adding in rulers to the renders offers consistent improvement. This could be attributed to the LLM getting more context of 3D positions of subjects guided by the rulers. We illustrate these options in the supplementary.

In Table~\ref{tab:sigmavsia}, we ablate over the identity-preserving image generation method. Since Insert-Anything can work with a single subject insertion at a time, each subject is inserted iteratively. We show that this leads to reduction in image quality which happens as in this method the image degrades on being processed consecutively multiple times. We also observe that Iterative Insert-Anything* leads to lower text alignment score, however, still better than other baselines.

\paragraph{Editing images in 3D}
Along with enabling better prompt adherence, our unique design of the spatial scratchpad combined with identity preserved image generation also facilitates robust, 3D controllable, and identity preserved image editing. In Figure~\ref{fig:editing}, we show both manual and text-based editing supported by our method.

\begin{figure}[h]
    \centering
\includegraphics[width=0.9\linewidth]{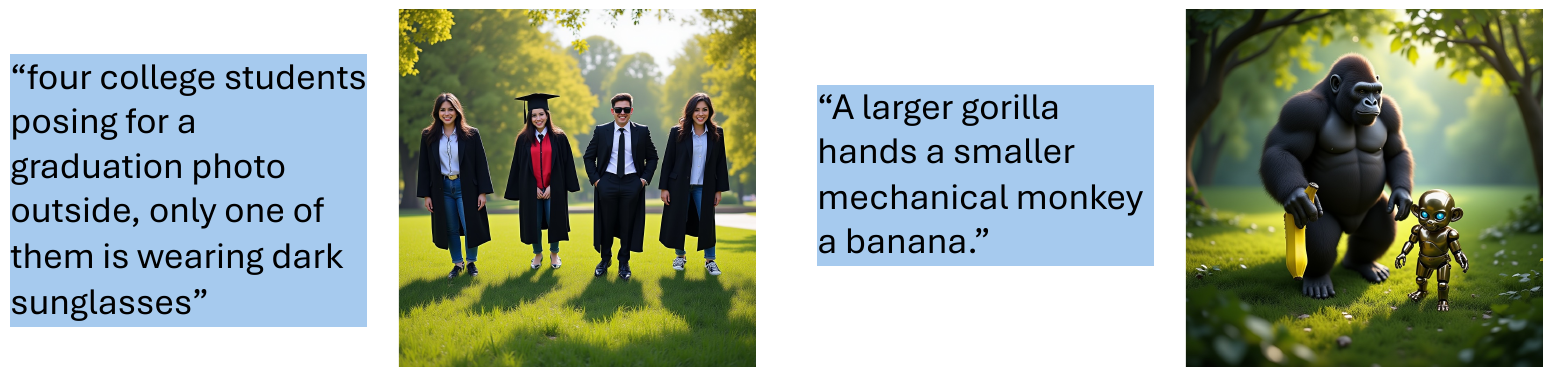}
    \caption{\textbf{Limitations.} We show that our LLM generated subject placements can be too uniform unlike real image compositions. Secondly, we find that activity and orientation understanding/planning may fail in cases of complex interactions.}
    \label{fig:limitations}
\end{figure}

\section{Conclusion}
We have introduced \textbf{3D spatial scratchpads} as a way to augment text-based reasoning in text-to-image generation frameworks. Beyond simple linguistic or 2D-based planning, our system enables direct reasoning of subject placement, scaling, and orientation in 3D using only existing LLMs.
Our scratchpad enables manual or text-based 3D edits in the generated images while preserving the identity
of the subjects involved.
Empirically, our method yields substantial gains on complex compositional image generation benchmarks, demonstrating the effectiveness of reasoning in a native 3D representation.

\paragraph{Limitations and future work}
As shown in Figure~\ref{fig:limitations}, our method has two main limitations.
In some cases, 3D arrangements created by LLMs may be too uniform and not reflect real-world compositions. 
Furthermore, generated 3D meshes may have limited pose diversity and our agents cannot repose articulated assets (\ie, the position and rotation of a deformable mesh may be adjusted, but individual joints cannot).
Future work may complement the pipeline with verification steps to improve composition naturalness, or incorporate physics simulators into the arranged environments to enable more plausible interactions.

\section{Acknowledgement}
OS and SM were supported in part by a NSF grant 2329927 and a Dissertation Fellowship from UMass Amherst.

%% file: sec/X_suppl.tex
\clearpage
\setcounter{page}{1}

\makeatletter
\renewcommand{\maketitlesupplementary}{
   \newpage

   \twocolumn[{ 
    \centering
    \Large
    \textbf{\thetitle}\\
    \vspace{0.5em}Supplementary Material \\
    \vspace{1.0em}
    
    \includegraphics[width=0.85\linewidth]{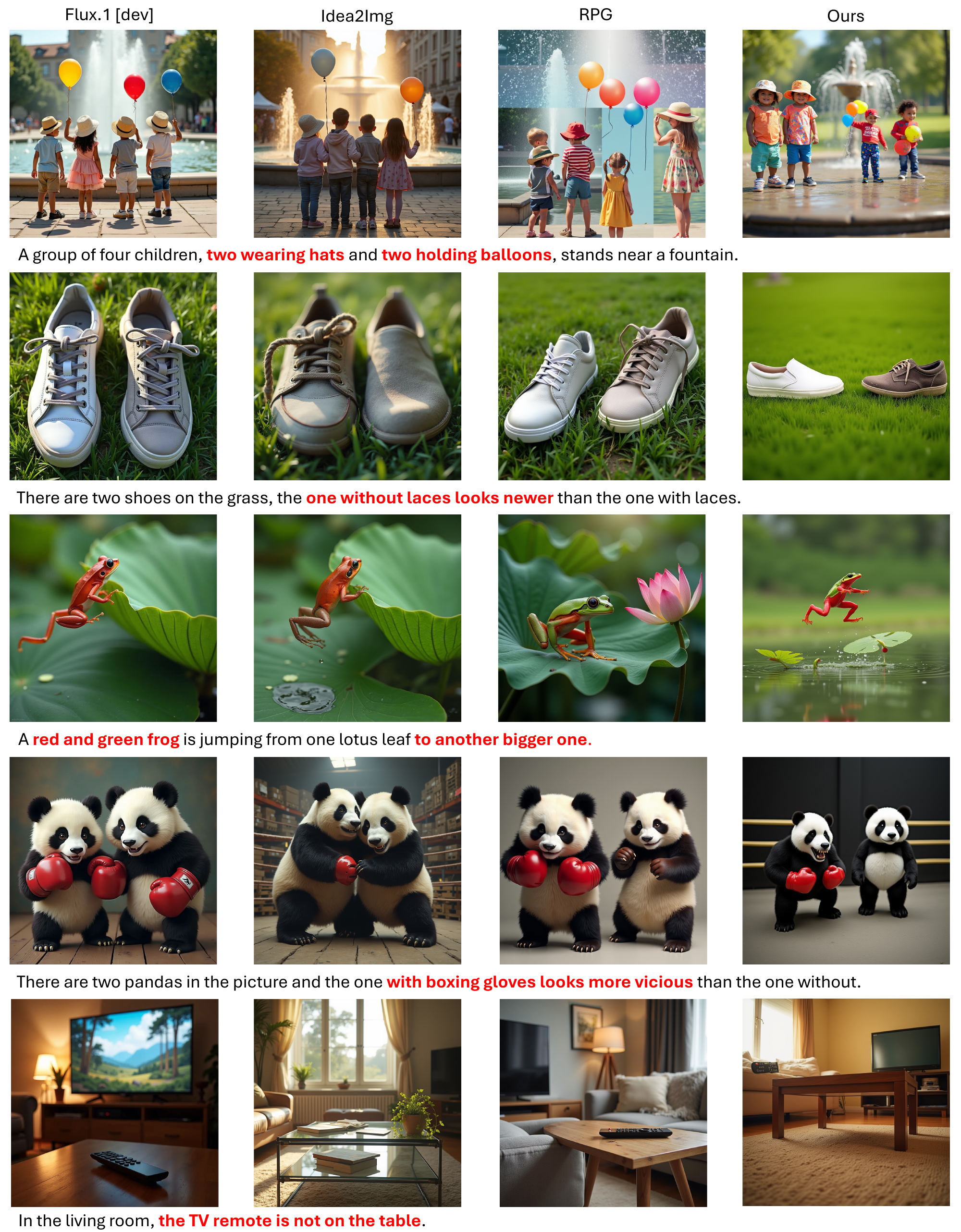}
    \captionof{figure}{\textbf{Additional qualitative examples from GenAIBench dataset.}}
    \label{fig:qual_supp_1}
    
    \vspace{1.0em}
   }]
}
\makeatother

\maketitlesupplementary

\begin{figure*}[]
    \centering
\includegraphics[width=0.85\linewidth]{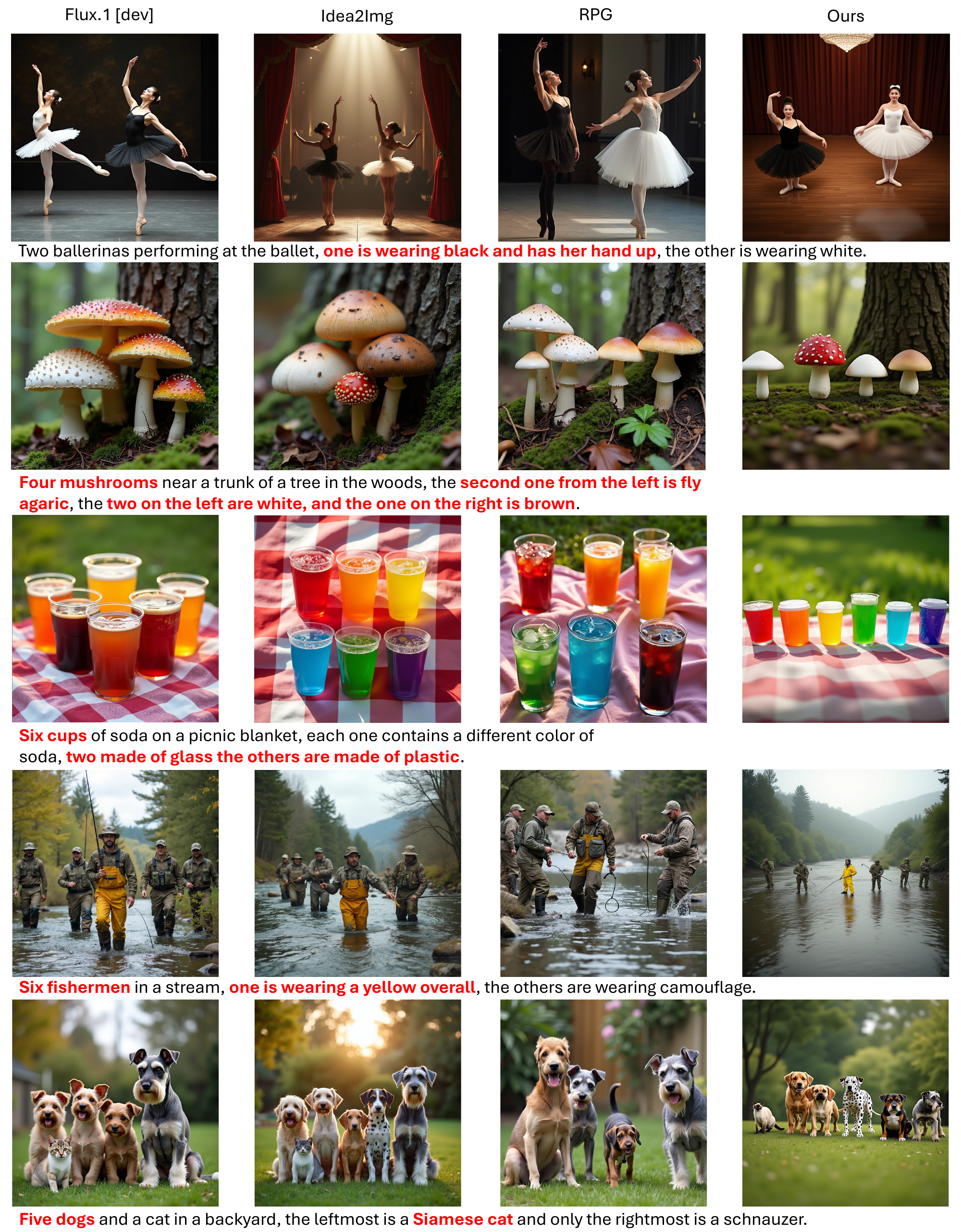}
    \caption{\textbf{Additional qualitative examples from CompoundPrompts dataset.}}
    \label{fig:qual_supp_2}
\end{figure*}

\section{Additional Qualitative Results}
We illustrate some additional qualitative results on the GenAIBench dataset in Figure~\ref{fig:qual_supp_1} and on the CompoundPrompts dataset in Figure~\ref{fig:qual_supp_2}. We highlight the errors made by the baselines with bold red text. These examples further show that our method excels in reasoning about counting, spatial, attribute, and comparative planning demanded by prompts for image generation.

\section{Reasoning about Orientation using LLMs}
\begin{figure*}[t]
    \centering
\includegraphics[width=0.85\linewidth]{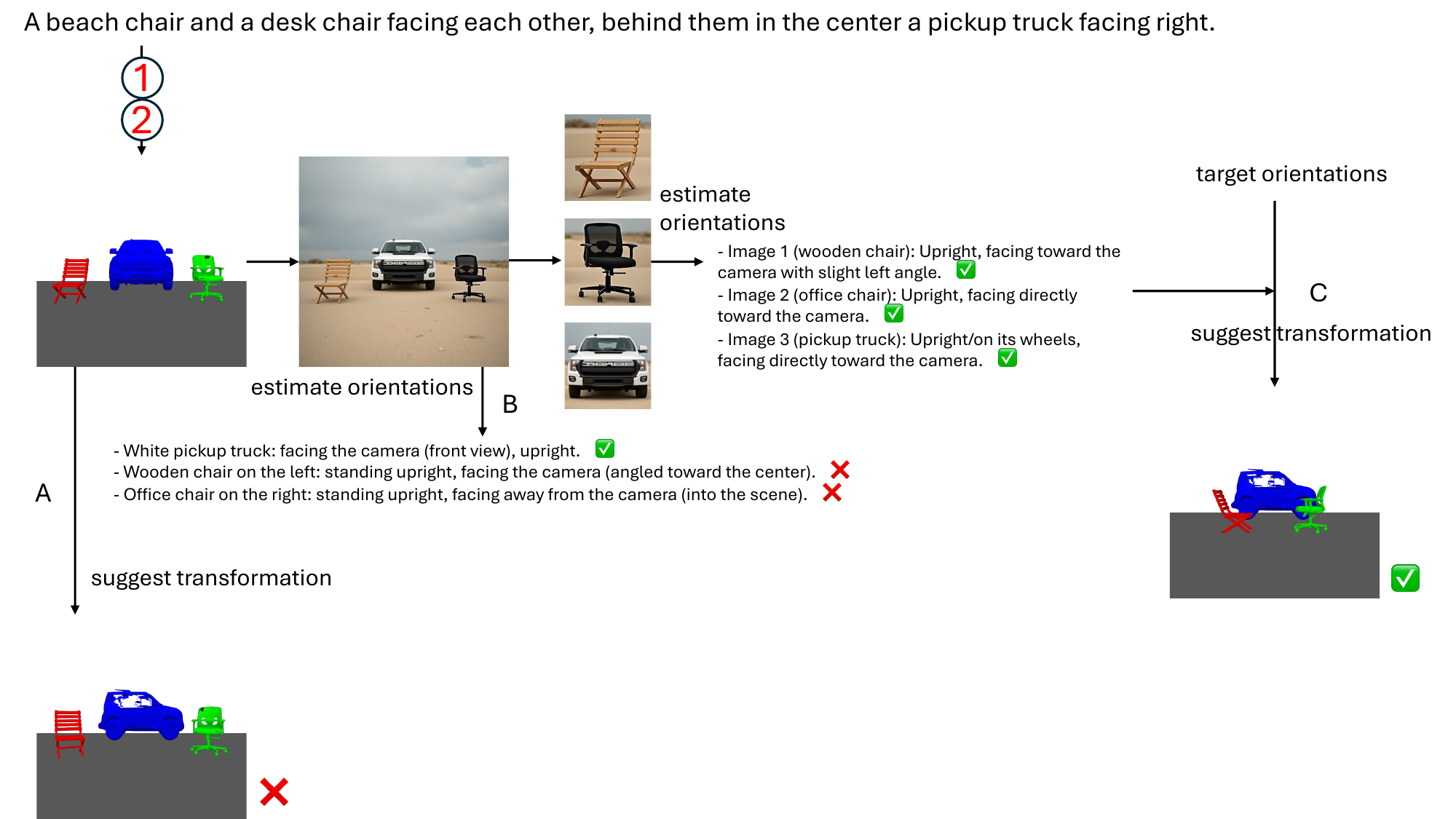}
    \caption{\textbf{Strategies for determining rotations.}
    We examine a scenario that requires accurate rotation planning for multiple objects described in a natural language prompt.
A: Directly providing the LLM with multiview renders of the entire scene and asking it to output transformation matrices fails to produce correct rotations.
B: Estimating orientations from the full synthesized image also proves unreliable: while the pickup truck is interpreted correctly, both chairs receive incorrect orientation predictions.
C: Our proposed strategy isolates each object by cropping its image, enabling the LLM to infer its orientation independently; these estimated orientations, paired with the desired target orientations, are then passed to a secondary agent that suggests appropriate transformations.
This crop-based decomposition yields far more accurate rotation predictions, improving robustness even though its quantitative impact is modest (Table~\ref{tab:agent_ablation}) due to benchmarks containing few orientation-sensitive prompts.
Nonetheless, the qualitative failures observed in (A) and (B) highlight the importance of dedicated orientation planning, motivating our explicit modeling of this capability.
    }
    \label{fig:orient}
\end{figure*}

In Figure~\ref{fig:orient} we explore a situation which requires specific rotation planning. We show that A - simply supplying the renders to the LLM and prompting to output transformation matrices does not work.
For estimating orientations, B - using the full image leads to erroneous estimations. Our strategy of C - using crops to estimate individual orientations and then supplying to another LLM along with target orientations is a more robust option. As shown in Table~\ref{tab:agent_ablation} the improvement offered by adding this agent is not significant due to benchmarks not containing many samples that require specific orientation planning. However, we focus on modeling this aspect robustly upon qualitatively noticing inconsistencies.

\begin{figure}[h]
    \centering
\includegraphics[width=0.85\linewidth]{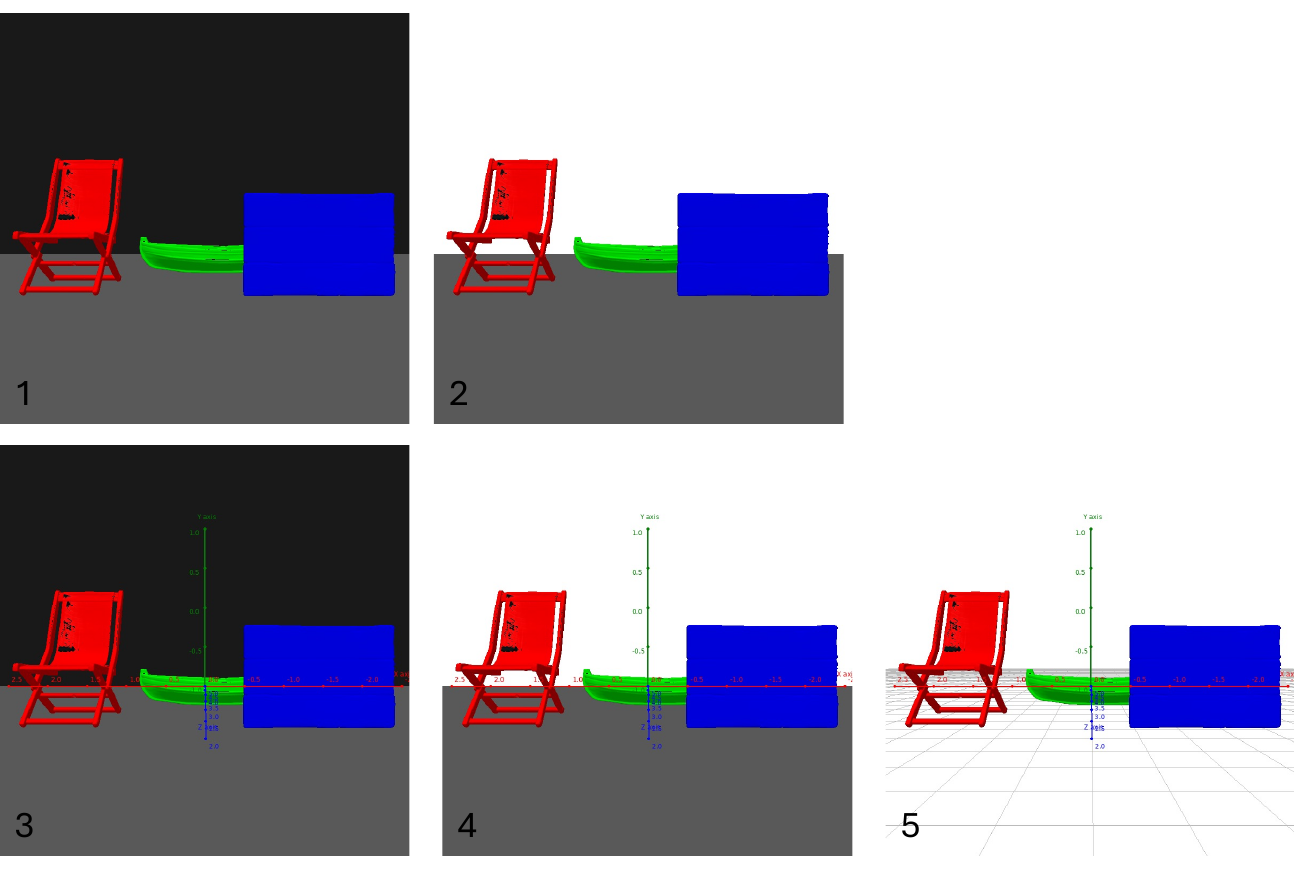}
    \caption{\textbf{Various rendering designs explored.}
    We illustrate the different rendering configurations explored in our design ablation, shown here only for the front-view camera but applied analogously to all viewpoints.
Each variant corresponds to the ordered settings listed in Table~\ref{tab:design_ablation}, including changes to background color, ruler visibility, and grid placement.
Rulers are drawn up to fixed spatial bounds to constrain object placement, and cameras are positioned with fixed viewing directions and distances chosen to ensure all meshes remain fully visible.
    }
    \label{fig:renderchoices}
\end{figure}

\section{Scratchpad rendering}

We show the various design choices we explore for rendering in Figure~\ref{fig:renderchoices}. We only show the front view here for each of the choices but the same strategy is applied for other views of each corresponding choice. The choices are shown in the order they were listed in Table~\ref{tab:design_ablation}. We draw rulers up till the fixed bounds seen in the figure so as to constrain placement within a region. The camera in each view is created with fixed directions per view (eg. $[0.0, 0.0, -1.0]$ for front) wrt to the center point of the scene and a distance which is fixed as the minimum distance required for all meshes to be completely visible. For rendering proposal views among which the final $\mathrm{CameraPicker}$ agent chooses the final view, we use the following directions:

\begin{verbatim}
    directions = {
        "proposal0": [0.0, 0.0, -1.0],
        "proposal1": [-0.3, 0.0, -1.0],
        "proposal2": [0.3, 0.0, -1.0],
        "proposal3": [0.0, 0.3, -1.0],
        "proposal4": [0.0, 1.0, 0.0],
    }
\end{verbatim}

These denote front, partial right, partial left, partial top, and top views in order.

\section{Evaluation on T2I-CompBench}
We evaluate the baselines and our method on 2D/3D spatial, complex, and numeracy tasks of T2I-CompBench. We use the same strategy as proposed by T2I-CompBench for evaluation, where for 2D/3D spatial tasks and numeracy we use UniDet~\cite{zhou2022simple} and for the complex category we use their proposed 3-in-1 metric. We observe that our method improves over baselines across all the reasoning domains. 

\begin{table}[h]
\centering
\scriptsize
\begin{tabular}{l|c|cccc}
\toprule
Method & \shortstack{Reasoning \\ modality} & 2D spatial & 3D spatial & Complex & Numeracy \\
\midrule
Flux       & -      & 0.29 & 0.39 & 0.37 & 0.61 \\
Idea2Img   & text   & 0.38 & 0.42 & 0.49 & 0.65 \\
RPG - Flux & 2D     & 0.43 & 0.43 & 0.55 & 0.64 \\
Ours       & 3D     & \textbf{0.48} & \textbf{0.54} & \textbf{0.65} & \textbf{0.67} \\
\bottomrule
\end{tabular}
\caption{\textbf{Evaluation on T2I-CompBench.}}
\label{tab:t2i}
\end{table}